# A Model of Spatial Thinking for Computational Intelligence

Kirill A. Sorudeykin,
*Relevance Research & Development Corp., IEEE Member,*
*Kharkov National University of Radio Electronics*
Kirill.A.Sorudeykin@ieee.org

*When the solution is simple, God is answering.*

*I want to know how God created this world. I am not interested in this or that phenomenon, in the spectrum of this or that element. I want to know His thoughts; the rest are details.*

*Albert Einstein*

## Abstract

*Trying to be effective (no matter who exactly and in what field) a person face the problem which inevitably destroys all our attempts to easily get to a desired goal. The problem is the existence of some insuperable barriers for our mind, anotherwords barriers for principles of thinking. They are our clue and main reason for research. Here we investigate these barriers and their features exposing the nature of mental process. We start from special structures which reflect the ways to define relations between objects. Then we came to realizing about what is the material our mind uses to build thoughts, to make conclusions, to understand, to form reasoning, etc. This can be called a mental dynamics. After this the nature of mental barriers on the required level of abstraction as well as the ways to pass through them became clear. We begin to understand why thinking flows in such a way, with such specifics and with such limitations we can observe in reality. This can help us to be more optimal.*

*At the final step we start to understand, what mathematical models can be applied to such a picture. We start to express our thoughts in a language of mathematics, developing an apparatus for our Spatial Theory of Mind, suitable to represent processes and infrastructure of thinking. We use abstract algebra and stay invariant in relation to the nature of objects.*

## 1. Introduction

Outward things demonstrate a lot of facts where interactions play leading role. Interactions can be found both in human activities and in nature. If we take a look at the science, we can find a great number of conformities. Chaos theory tries to predict a position and behavior of small particles in a wide area of space after a relatively long period of time. To do this it uses Probability Theory. Quantum mechanics tries to describe a nature of fundamental forces of physics such as weak nuclear interaction. The study of gravity, which is based principally on the General Theory of Relativity, describes interactions of macro objects like planets. Electromagnetism, gravity, weak and strong nuclear forces should be united in the Theory of Everything or in another words a Unified Theory. This is one of the most important courses of modern theoretical physics during more than a hundred years. And it is still unresolved. There are several candidates for the solution: Field Theory, based on the differential geometry, Strings Theory which rides on the supersymmetry, an Exceptionally Simple Theory of Everything, grounded on the Lie algebra. But all of them endure hardships now.

Unified Theory is not a prerogative exactly of theoretical physics. In another branch of science, economics, Games Theory is widely used. It completely relies on the specificities of objects' interactions. Non-cooperative games say that we need a general view on the system dynamics to build optimal strategies for each separate object. For that reason interactions between objects within the bounds of the mentioned system will have a high degree of mutual organization. The same process we know as a Synergy. This research branch is concentrated on the phenomena of self-organization and autowave processes. As well as Darwinian Evolution it studies interactions within systems. Natural selection provides a succession of improvement steps which gradually adapt objects to the dynamically changing environment. Emergent behavior illuminates the nature of irreducible complexity that is the effect of cardinal changes in systems' structures. Another example is a phenomenon of self-similarity widely represented in nature and being studied by means of mathematical fractals.

Intellectual activity is full of interactions too. Mentality is merely interactions between imaginary objects in a virtual space which we call mind. But the latter is more like a "battlefield". Trying to make crucial decisions, you take a risk of entering into the state of stagnation, when you become unable to choose the better or sufficient solution to your problem. This is a huge impediment to creative work. However, a succession of turning points can limit a number of possible alternatives, simplifying a choice. Known as complexity, this problem is a foundation of Computer Science. Also it is the departing point of our theory. As for the mental discipline, you need certain external influence, support, i.e. a backbone to produce optimal solutions. It helps to sight most important correlations in the subject of action. For example, design patterns are useful in constructing effective software architectures. Also strong will is necessary to direct your thoughts and concentrate your mental energy. Anyway you must continuously move forward to the achievement of your goal because of this is only way to select specific path of further developments among multitude of ones.

## 2. Problems Statement, Making a Choice

Knowing "why" gives us an ability to judge about "how". Having specific task we start to analyse what is located "near" it. And this includes our own experience, our thoughts, images and ideas, all the data and all the information we ever saw, heard, felt, and everything we can remember and relate to this. In other words, we study a context of a problem.

Then we start to make "**leaps**" from- and to another notions, fields of knowledge, etc. to extend our vision even more. We became able to think about the causes and effects, to connect something previously unknown, to find a sense in things we meet. All these allow us to "recognize" reality: when we meet something new we start to look for a departing point to understand it. And it is a departing point that constitutes the substance of our theory – what is it, how it works, what is it consists of, etc. We need a departing point because…

Because we need to know the origin of notions or their connections to something already known by us. I.e. we need to orientate ourselves in a space that we see, to get a role of each object in our space. When we meet something new we cannot grasp it at the moment, but need to survey it first. We need to switch context, to digress. This means that the content of things lies in the depths, not on a surface.

The scheme below demonstrates a typical process of thinking when we seek a solution to some problem. Analyzing any more- or less complex task we come from three shapes: $1^{st}$- a task description, $2^{nd}$ - our own experience and knowledge and $3^{rd}$ - data from surrounding world, taken by experiment or by communication with other people or objects (Figure 1). First we try to use our experience to recognize initial task. We build a virtual model of it in our mind, representing a subject domain and finding correspondences between our view and initial task description. We try to superpose initial task description with our initial view of solution, determining the difference between them. Seeing the difference, we can determine the steps which allow us to come from our starting vision to the desired or interim solution of a problem.

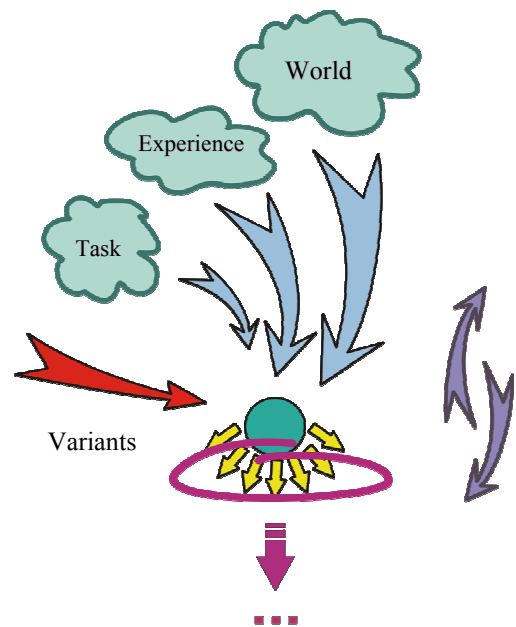

**Figure 1. A process of reasoning**

But some time later we face the problem of a contradiction, when we cannot choose any variant from arisen alternatives, because our experience is appeared not enough to give the answer on the question of what variant is more optimal and best suitable for our task. I.e. we see the problem that we mentioned above – we need some knowledge about objects we deal with before we model some situation. So, we apply external knowledge – from real experiments, from some knowledge sources and with this information we can move forward. Such actions are repeated again and again right until we meet the same contradiction, as before. But this time even external knowledge together with personal experience is not enough to make a decision.

What is the nature of this problem? Variants are generated as a set of assumptions, because we don't know direct answer yet. We cannot differentiate these variants, as we don't see any advantages of one solution before the other. To determine difference we need to conduct an experiment, but it is possible that in specific situation we will no longer have a possibility to check, just to make actual solution. I.e. we need an answer before the trial.

As an example of such a situation we can consider a problem, which I call "a paradox of paper". This is a typical situation for every person who dealt with writing somewhen. Let us suppose that we need to write a text describing some idea. As soon as we take a sheet, we begin to lose our idea. This happen because accomplishing a physical action needs us to switch attention to this at least for a moment and temporary lost our focus from the previous point. Then, after a beginning of writing, we partly recover the information in our mind and encounter another problem – our idea has more ideological (constructive) lines than sheet of paper able to "take" from us in a single moment of time. We write linearly while think with images which are not linear. Sometimes we need to express several things simultaneously, but able to express just one of them in a time. Trying to express one point, we can lost a whole picture, which makes impossible to continue explanation, and require us to recover main idea again. We need to repeat constantly some points, phrases and words in order to put together separate parts of whole construction which we have been forced to divide earlier. We need to decide in what order these parts should be represented in a text. Gradually original idea undergoes a considerable change in order to be expressed by means of paper. Some words, that we use, cannot describe exactly that sense what we need and we have to spend hours to find other words, but anyway they will not be absolutely perfect for us.

What to say, it seems like unsolvable task - to avoid these hardships. But people invented a solution a long time ago – a language, a speech. Of course, language, especially writing, is not an ideal thing, but if we don't know how to call something, we are always free to invent a name for this. Since earliest times people did that and keep expressing complex thoughts by just usual words for now, inventing new terms. Again, we had a problem of explaining some complex ideas by some vocabulary. If this vocabulary does not fit to the idea, we modify vocabulary, i.e. we add something new to already existent. We tried to find a solution and when this gave no satisfactory result, we modified the environment.

Each variant is characterized by a set of properties by which we can compare them. If we add new information to a space of choice, then this set will be modified too.

## 3. Context Thinking

Words and notions, related by meaning, specific points of view, specific objects and specific incidents, definite moments of time… Everything is gathering around the idea of concreteness (or distinctness), which in turn is the basis for occurrence of discreteness. We already mentioned a term 'context' earlier in this paper. Schematically, context can be represented as on the Figure 2. Formally, this is a representation of notion 'ordering'. Let's consider a mechanism (or phenomenon) of context thinking in more detail. Especially here we come from the point that this is the only (or at least the main) principle of thinking and way of organizing a mental space.

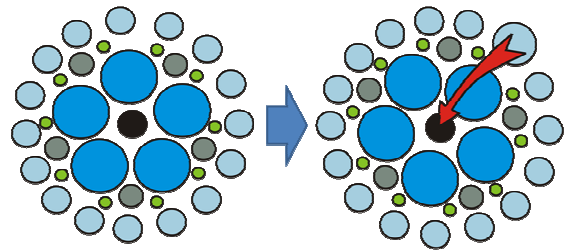

**Figure 2. A scheme of context thinking**

The problem of representing context thinking is that we need a specific mathematical tool to show neighborhood of objects of different rank. Saying more detailed, we need to know, what objects should be neighbors, what objects should be between neighbor objects, how to differ a rank of objects (in such a way we can identify, what objects or points will be reference and what interim). Then another question is how points should connect each other. What criteria or interface should be used for that purpose? And what criteria will be missed? If some criteria or types of connections will be missed, this will constrain the number of objects (or types of objects) which can make direct connection with some current object. This in turn leads to appearing an extent of a space in some specific direction or sense. This creates space limitations and appearance of time as an exhaustible resource.

In such an apparatus we need to decide – does anything should be in a center of such a context? I.e. does a central (reference) point necessary for such a structure, or not? The root of this question is that if yes, then what object should be selected as the main and by what criteria? If no, then how to, actually, form a context, around which? As you see, we met a dilemma here, and there are a lot of other such dilemmas in our subject domain, which we are definitely required to solve.

Speaking about *space limitations* [11], we need to conclude that they are defined by the rules of space itself. Here we again come to important reasoning about space completeness – time appears as a result of limitation (imperfection) of space, which cannot absorb everything simultaneously and hence, needs to include things in time and in extent, forming what we see as a usual space. This way ordering appears as a natural property of space.

## 4. Mental Process

So, thinking appears as a complex of different processes inherently based on the phenomena of ordering. Understanding of things can be represented as finding places for them in a system of individual view, i.e. finding a specific order position for each object by using a procedure of comparing. To understand something we need to put it to our view, sometimes changing one, extending it, and then start to compare objects. By such a way we can determine priorities of things, categories, ideas one in relation to other. If we don't have a prioritized view, we became disorientated, i.e. unable to select a specific direction of thinking. That's why we naturally need order.

This chapter is a good basis for starting to speak about mathematical apparatus. From here we can come to any task related to the mechanisms of thinking, such as optimization and modifying, automated design, sorting, decisions making, inventing, recognition (classification), feature extraction, etc. This basic mathematical apparatus is built with mathematical formulation of some specific patterns of thinking. We meet them in everyday life, but to get their nature, we need to reason about them and find strict formulations.

Below there are several examples.

### 4.1 Examples of elementary mental processes

- Each person has unique experience. Everybody knows that. But you can have no the least suspicion that people which surround you every day, which meet you so often on the streets, at work, at home, even did not have any idea about some things that are absolutely natural to you. Sometimes forms of difference between outlooks can be totally nonpresumable, for example under the same words we can imply completely different meanings. And this often leads to various even most surprising forms of misunderstanding.

- When you got a goal and start to move to it, you begin to throw off any variants not related to the goal. This means that you understand the goal, the direction to move to and the "space" that you see in your mind is built with an extensive structure. But we also can call this a preperception, i.e. an ability to feel something – good solutions, right direction in a mental space, etc. Is it possible that we really feel these solutions by someway? In that case how this mechanism is realized?

- When you look onto a specific task, let it be a software product, picture, wooden construction or something else, you are trying to take into consideration everything that can be necessary, every detail. Then you sort this data by someway and build a specific algorithm to accomplish your goal. Some variants, that you are in principle able to suppose, will be not applicable (or improbable) to your task from the very beginning. You at once determine this and will not include them to your plan, which is more effective.

- If you try to do something in your new field, you feel difficulties because of psychological pressure. But as soon as you started, acting became easier. And if somebody supports you at the time of start, you can even start easily too and will not have any problems.

- If you decided to imagine *everything*, you can imagine just everything *you know*. But no one can say that this is completely everything. Always there is something more, even if you know nothing about it yet. Because of any knowledge cannot be absolute.

- When you imagine something and then try to detail certain part of your idea you lose general point. At once general thought can be so huge, that you cannot completely formulate it in a time. You just feel it by some way. Transient ideas about some parts of general idea came so suddenly and disappear so quickly that you need to catch them and fix by some way as soon as possible in order to keep it, to prevent its losing, to record this information and your feelings about it. Then you will be able to restore this information and depart from it developing and deeping your view.

- You need a departing point to reflect on specific things. Mind strives to something whole, fine, defined. Attention should be always riveted on some objects, some bright images. To formulate something you need to deepen into this. And until you find a strict point you will not be able to express your idea (how to do this is another topic). We can say that from this point of view mind strives to concreteness. If you would like to understand something solid, i.e. be aware of it, you need to imagine clearly a position of each new object in relation to others in the space of mind.

- Different people able come to the similar conclusions and same results acting completely

independently. We even able to feel harmony and strive to it in our thoughts and actions.

- Automation of actions: our individual style, individual behavior, habits – anything that can help us to be more effective. This forms a kind of experience which we able to train. Some actions become steady modes (patterns) and we act our usual way preferring it to some other possible ways.
- Sometimes it is easier to "jump" between different notions than connect them with a single reasoning line. This can make a process of explaining quicker. And maybe any reasoning on some level of abstraction consists of such discrete leaps.
- It is very hard to simultaneously control and act. These are antagonistic tasks – control requires us to keep whole picture in mind, but acting needs to go into detail of specific work.
- When you cannot choose anything a feeling appears that something holds you all round, not allowing a move to any direction. And to move you feel like you need to "break this chain" and allow only one mounting to remain; all others need to be removed.
- It is very easy to memorize relations between limited number of neighbor objects in some space than keep in memory all existent objects and their relations. By such a way we can get reasoning involved into a process of remembering, making it more effective. Instead of remembering all the information, you can remember just local details and restore general point by thinking. This reduces the amount of information necessary to represent things and shortens the procedure of searching. By the way if we need to generate new knowledge this approach also works.
- There are some special strategies of thinking which can more effectively bring a person to result allowing them to keep mind in order and be in a good shape to make decisions. They can be called Mechanisms of Mental Effectiveness. How to determine them? All in all the same way as mental space is designed. As we need to achieve some results we need to know, how to do this, i.e. results should be planned. This means that some structure should underlie the actions and we should follow this structure. We need to analyze task deeply and propose a comprehensive solution to it. Then *main direction* should be kept no matter what external influences we are experiencing, because of our vision of solution needs to be sufficiently inertial to allow us to get to the goal. It is just necessary to take notice of all the required details and take advantage of them to implement our strategy. In this process independency of points of view is very important (maybe this is the basic element for mental effectiveness), because we need to care about stability of the strategy, which can become blurred if we lose concentration. So, we need to be plunged into our vision, taking care about its clearness as well as trying to avoid any mental contradictions and meshing. We need to keep position integral which should allow us to develop activity and move forward with it, otherwise mental process will not be as effective as possible. As we see, mental effectiveness to a large extent is a disciplinary task, which depends on the ordering of information in a mental space together with the element of unexpectedness based on the individuality of different points of view.

These and many other stamps make process of thinking a sum of typical elements, a part of something bigger and maybe more fundamental but at once more simpler because we can see here some degree of regularity. It demonstrates a principal property which on our sight characterizes a mental process – **impartial** relation to objects from the point of the space. All the objects are in some degree equal, because all of them are children of the basic space. All they have equal chances to be everywhere and specific place of single object depends basically on the behavior of it, which determines the structure of space, while the latter itself has more inertial laws. On the basis of this property we can build a very interesting and very universal theory, able also to represent our vision of mind.

**4.2 Examples of mathematical notations**

One of possible ways to solve a task of choice is an exhaustive search. You should check each variant and successively find a better one. But people have another way – intuition or preperception. Is it really possible to select right variant at once, without any seek? We will try to answer here.

A scheme of the task is represented on the Figure 3. Any choice, if we make it by an exhaustive search, will need a time, proportional to the number of variants. If we select a solution quicker, this will give us ability to increase our productivity in variety of applied problems and change our views on the Theory of Complexity. Really, to solve this problem momently without a search we need to have a previously prepared database, where we already have an answer. We come to realizing that we need to change our view on the complexity to represent it in more suitable way. Any process of seeking a solution – is just a process of learning, nothing more. Answering means using the experience to give the answer. The process of search and giving the answer lie on the different levels of abstraction from the point of view of complexity and should not be mixed together. They should be mathematically separated. During the process of thinking we represent a task in a way, which allow finding the answer easily. Anotherwords, we set problems and this allows us to "see" the answer like people do using their intuition.

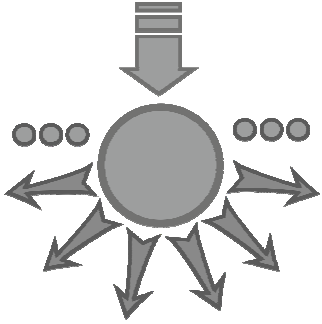

**Figure 3. A problem of choice**

Consider that we need to make a choice from some number of alternatives. The task will be in the center and possible solutions will depart from it. Let's take following definitions:

$T = \{t_m\}$ – source task with their components;

$V = \{v_n\}$ – collection of alternative variants;

$A = \{a_l\}$ – action algorithm with steps;

$m, n, l \in \mathbb{N}$ - powers of sets

P – new defined point in a space.
D – distinctness.

We have a family of functions:

$$F = \{f_j(S)\}, j \in \mathbb{N}$$

Each such function translates a number of variants to one single variant and hence to a collection of further actions (algorithm):

$$D = f_j : V \to (A \parallel P);$$

Functions of F can be various and specific of them can be chosen dependently on the situation. But universal function $f_{un} \in F$ represents the following.

Facing the problem of $V \to (A \parallel P)$ we begin to collect information about the subject domain to be able to make a choice:

$$S \to S \bigcup K_i(T); i \in \mathbb{N},$$

where $K_i$ - elementary portion of new knowledge, S – space of subject domain, $o_s$ - object in S:

$$S = \{o_s(S)\}, s \in |S|$$

I.e. space S is combined from objects, related to each other, making a specific order.

That's why collecting new information we obtain each time a modification of set S:

$$S \to S'$$

We use here a special notation $o_s(S)$ or $K_i(T)$ to show relations between constituents of these expressions. They represent a direction of thinking, for example "how $o_s$ is related to S", etc.

We should note that $o_s(S)$ means here a ε-neighborhood of specific point $o_s \in S$:

$$U_\varepsilon(o_s) = \{o \in S \mid \delta(o_s, o) < \varepsilon\}, \varepsilon \to 0$$

Otherwise it will be

$$t(o_s(S)) \to \infty, S \to \varnothing$$

(t – function of time), which is inappropriate.

$\delta(o_s, o)$ denotes a metric of this space, i.e. how to calculate distance in there. We will discuss it later.

Then it should be true that already $[T, V, P] \in S$. In result our task of function $f_{un}$ transforms into this: (For some other goals any other function can be used, for example, classifiers, evolutionary and genetic algorithms, etc.)

$$f_{un}(S) = \Theta_S(dP, gP)$$

where $\Theta$ - *transitive closure* between given points in a space S, dP – departing point, gP – goal point.

$$dP = T, gP = P, P \in A.$$

(according to our earlier definitions)

Space S already contains a solution to our problem after collecting information about subject domain and we don't need to make a choice at all. We need just to select appropriate points and connect them right. Saying more, knowledgebase for automated design should contain a system of vectors representing right connections between notions:

$$DB = \{\Theta(dP, gP) \mid dP, gP \in P(DB)\}$$

In order to allow interactions between objects they should be represented in a single common space E:

$$E = \{DB_c\}; \Rightarrow iO[DB_a \rangle \langle DB_b] \langle f: E \to DB \rangle$$

iO[x]<y> quantifier means "in order to *x* do *y*".
Symbol "><" means abstract interaction.

The latter expression represents a ratio which we call "**necessity**". As interaction already taking place, there is necessary that both objects are expressed with a single system of terms (future in the past).

## 5. The Notation of Discreteness

Let's start from the primitive. On the Figure 4 you can see the schematic representation of what we call a succession of steps, i.e. simple coordinate axis with a number of regular countings. If we look on the picture, we can see actually the principle of additive measure. A bottom image of picture represents another interesting feature of successions – a *space limitation*. This is the most basic term. We can't get to the next not passing through previous. This is obligatory property of *mental tunneling* and any successive phenomena. (Some sources call it linear-dependency, etc.)

Let's call our algorithm (some action, behavioral graph, etc.) a Space S. The structure of it should be described like this:

$$S = \{s_i\}, i = \overline{0..n}$$

$s_i$ – Steps (countings on some axis) of some succession of actions (algorithm). n = |S|

In general,

$$s_i = f_i(s_{i-1}), f_i(s_{i-k} \mid k > 1, k \in \mathbb{N}) \notin S$$

i.e. linear structure gives us a dependency between current and previous step as a function $f_i(...)$.

Let $T_i = \{t_{ij}\}, j = \overline{0..m}$ be a Feature Set (i.e. transition vector) of characters used to move from previous step to the next. Generally $T_i \neq T_{i-k}$ i.e. transition vectors are individual for each step. Of course, some of them can coincide, but departing point is that they do not match. In effect vector $T = \{T_i\}$ represents an individual axis in a coordinate vector space of some object model which we have in reality. This is a partially ordered set. We obtain $S = T$ as the basis equation, and hence $f_i(s_{i-1}) = T_i$. Most simple case is when we have just singular feature set for all the counting on an axis. We have such case in Euclidean coordinate system.

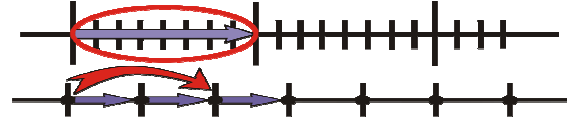

**Figure 4. Discreteness and limitations**

Then we have the relation of superiority: $s_i < s_{i+1} < s_{i+2} < ...$ with some sense as the main property of coordinate space (axis). As we see, here we don't have a direct connection between $s_{i-2}, s_i, s_{i+2}$, etc., because of which we cannot directly compare them. That's why we cannot say what actual relations these points have between each other. We can only suppose that the same as between neighbor points. We have bottlenecks of $s_{i-1}$ $s_{i+1}$ which we should go through in order to move to the next step. Why superiority? Just due to the ordering nature of linear space. We have a queue of steps here and each step has its own personal number (position). There is no other way to dispose objects in such a type of space. Each object can be represented as an intersection of certain number of axes and in cross-point we have new notion, i.e. our object O:

$$O = \bigcap_l S_l; l \leq L \in \mathbb{N},$$

where *l* means an *l*-th axis of total *L* of them.

Each axis represents some individual aspect of the object and particularly near the cross point represents a composite part of an object. The following expression should display a composite parts $o_m$ of object, united by a XOR operation:

$$\{o_m\} = \otimes_l S_l; m \in \mathbb{N}$$

Practically, the equation $O = \{o_m\}$ means not only definition of the set O, but also the operation of Synergy, when from separated parts we are obtaining a new category. The full view of the space in that case will look like this operation:

$$M_o = \bigcup_l S_l ,$$

where $M_o$ is a Map of object O. It might be useful if we, for example, trying to investigate a hierarchical structure of space of solution.

We can represent this operation also by the new notion "Abstract Integral", which should mean the general view on the *family* of L curves:

$$M_0 = A \int_{l=1}^{L} S_l \, d(l)$$

## 6. Practical Applications

Demonstrated technique of mathematical description of tasks, which I call "Representing of Reality" can be used in various applications beginning from calculating optimal strategies and completing with automated design of engineering systems. Particularly here you can see the example of a task of creating an optimal strategy in intense environment (Figure 5).

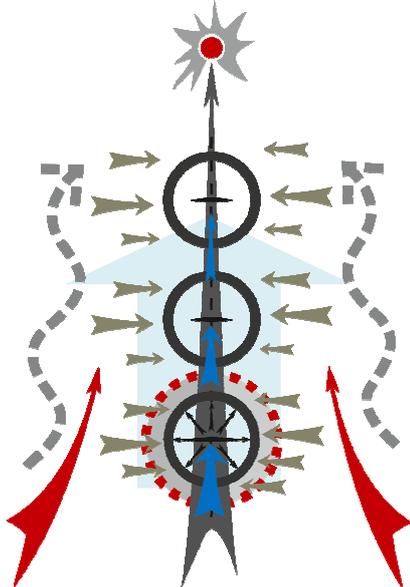

**Figure 5. The nature of contextual structure**

This model demonstrates the outstripping as the main strategy of control. Context moves forward each time before any surrounding dynamics had time to influence on the strategy. That's why, in correspondence of our mathematical apparatus, the strategy can use external dynamics to make steps further, staying oriented onto the same goals and being inertial in that sense, which increase productivity.

## 7. Conclusion

We considered a theory of mind. To be true, this is not a simple theory. But we will explain it step by step process for process to throw a light on the principles of thinking and to allow this apparatus to represent reality as deep as possible for mathematical tool.

You have been introduced to a method we use to investigate. You got to know about the tasks which we are interested in analyzing. You saw the basic points of our theory. Further works will show you another sides of it and we will develop our view in a most deep way.

Our goal in this research is to get into the mechanisms, or laws, which allow us to be so productive, creative and effective in imaginative tasks. We should explain this and, believe, it will help us to understand the idea of Creator in more detail, which will open us the way to the wisdom of nature which we still were unable to learn. Maybe it can help us to know ourselves better and to organize our lives in more profound way, allowing harmony to be a leading force in various relations people got into since the earliest times of existence of civilization.

## 8. References

[1] C. Ghezzi, M. Jazayeri, D. Mandrioli, "Fundamentals of Software Engineering", PrenticeHall, 2003

[2] Jack Trout, Steve Rivkin. "The power of simplicity", New York. McGraw-Hill. Nov. 1998

[3] Bondarenko, Hahanov, Guz, Shabanov-Kushnarenko Infrastructure of brain-like computing processes. Kharkov, KhNURE 2010. – 160 p.

[4] Kirill Sorudeykin. "A Research Methodology of Systems Design Nature", Proceedings of Third International Conference "Applied Electronics: State & Perspectives", Kharkov, Ukraine, 2008

[5] Kirill A. Sorudeykin. "An educative Brain-Computer interface", proceedings of IEEE East-West Design&Test International Symposium, Moscow, Russia, September 2009

[6] Kirill Sorudeykin. "An Analysis of Mechanisms of Economical Synergetics". Proceedings of 12 International Conference "Radio Electronics in 21 Century", Kharkov, Ukraine, 2008

[7] George F. Luger. Artificial Intelligence: Structures and Strategies for Complex Problem Solving. 5th Edition.

[8] Hamdy A. Taha. Operations Research: An Introduction. 8th Edition.

[9] Albert Einstein. Collected Scientific Papers, Volume 4, "Science" Press, Moscow, Russia, 1967

[10] D.I. Kryzhanovsky. Parametric identification of non-linear dependencies using simulated annealing algorithm. // Proceedings of the conference "Microsoft technologies in theory and practice of programming" Nizhny Novgorod, Russia, NNGU, 2009. – PP. 220 – 225.

[11] Kirill Sorudeykin. "An Operational Analysis and the Degree of Inertia in the Modeling of Thinking Process". Proceedings of VI-th international conference "Science And Social Problems of Society: Informatization and Informational Technologies", KhNURE, 2011, PP. 335-336

[12] Kirill Sorudeykin. "Levels of Ordering in Software Design and Thinking". Proceedings of IV-th all-Ukraining conference "Intellectual Computer Systems And Networks", Krivoy Rog, Ukraine, 2011, PP. 178-181